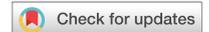

OPEN

# Quantum inspired qubit qutrit neural networks for real time financial forecasting

Kanishk Bakshi & Kathiravan Srinivasan✉

This research investigates the performance and efficacy of machine learning models in stock prediction, comparing Artificial Neural Networks (ANNs), Quantum Qubit-based Neural Networks (QQBNs), and Quantum Qutrit-based Neural Networks (QQTNs). By outlining methodologies, architectures, and training procedures, the study highlights significant differences in training times and performance metrics across models. While all models demonstrate robust accuracies above 70%, the Quantum Qutrit-based Neural Network consistently outperforms with advantages in risk-adjusted returns, measured by the Sharpe ratio, greater consistency in prediction quality through the Information Coefficient, and enhanced robustness under varying market conditions. The QQTN not only surpasses its classical and qubit-based counterparts in multiple quantitative and qualitative metrics but also achieves comparable performance with significantly reduced training times. These results showcase the promising prospects of Quantum Qutrit-based Neural Networks in practical financial applications, where real-time processing is critical. By achieving superior accuracy, efficiency, and adaptability, the proposed models underscore the transformative potential of quantum-inspired approaches, paving the way for their integration into computationally intensive fields.

Quantum computing represents a paradigm shift in computational capacity, leveraging quantum mechanics to achieve unparalleled processing power and efficiency. Within artificial intelligence (AI) and machine learning (ML), this quantum framework enables the development of advanced quantum-inspired models, specifically Quantum Qubit-based and Quantum Qutrit-based Neural Networks (NNs), which are the focal point of this study.

Quantum intelligence, situated at the intersection of quantum computing and AI, harnesses phenomena like superposition, entanglement, and interference to optimize and solve intricate prediction tasks that classical methods struggle to address effectively. This study fills a notable research gap by empirically investigating Qutrit-based NNs, which remain underexplored despite their potential for enhanced data representation and computational performance[1,2].

While theoretical applications of quantum computing have been extensively examined, practical implementations especially within predictive tasks such as stock market forecasting are scarce. This research employs stock market prediction as an empirical testbed to examine the effectiveness and scalability of Qubit- and Qutrit-based quantum models. Beyond market prediction, this work offers a foundational analysis of the broader implications of quantum intelligence for complex computational tasks[3–5].

Recent advancements in hybrid quantum neural networks (QNNs) have underscored the practical potential of quantum-inspired approaches in financial forecasting. Paquet and Soleymani[6] introduced the QuantumLeap system–a hybrid deep quantum neural network that employs density matrix transformations within a deep quantum architecture–to enhance financial time series predictions, particularly in regression and extrapolation tasks. Similarly, Gandhudi et al[7] developed an explainable hybrid QNN to evaluate the influence of social media sentiment on stock prices, thereby addressing the interpretability issues of traditional black-box models. In addition, Liu and Ma[8] proposed a quantum artificial neural network inspired by the Elman neural network, utilizing quantum genetic algorithms for optimal learning rate adjustment, which further substantiates the efficacy of quantum-enhanced architectures in capturing market dynamics. Building on these seminal studies, our work extends the quantum financial forecasting landscape by investigating Qutrit-based neural networks–an underexplored domain that promises superior computational efficiency and improved data representation.

School of Computer Science and Engineering, Vellore Institute of Technology, Vellore 632014, India. ✉email: kathiravan.srinivasan@vit.ac.in

Recent advances in gate-model quantum neural networks have established robust frameworks for implementing learning procedures on near-term quantum devices. These studies demonstrate that by employing constraint-based training optimization and unsupervised control of quantum gate operations[9], one can significantly enhance the performance, Further by reducing circuit depth through optimized unitary transformations[10], one can significantly enhance the performance and convergence of quantum neural architectures[11]. Innovations in circuit depth reduction and the optimization of local unitary operations further refine these systems, ensuring both computational efficiency and stability on gate-model quantum computers. In this context, our work on Qutrit-based neural networks not only complements these developments but also opens avenues for integrating quantum-inspired models within established gate-model frameworks[12].

In essence, this research enriches the field of quantum intelligence, illustrating the pragmatic potential of quantum-inspired neural networks within real-world scenarios[13,14]. Through rigorous analysis, this study contributes to advancing quantum computing's integration into AI, forecasting transformative progress across computational sciences.

Key contributions include:

(a) Demonstration of robust model performance (>=70%) in stock market prediction, confirming model generalizability across data sets.
(b) Exploration of hybrid quantum NNs, addressing an identified gap within the AI-quantum landscape.
(c) Comparative analysis of Qubit-, Qutrit-, and classical NNs, providing insight into each model's strengths and constraints.

## Methodology
### Classical bit neural network

Artificial neural network (ANN) was designed with key activation functions, cost function choices, and forward and backward propagation, optimized for application in binary and continuous prediction tasks. In this manuscript, the terminology *Artificial Neural Network (ANN)* specifically refers to a conventional, classical bit-based neural network implemented on traditional digital hardware.

*Activation functions*

Activation functions introduce essential non-linearity, determining whether neurons propagate outputs across layers. Standard choices like sigmoid, tanh, ReLU (Rectified Linear Unit), and softmax govern the flow and complexity of information. For an input $Z$, the sigmoid function is given by:

$$\text{sigmoid}(Z) = \frac{1}{1 + e^{-Z}} \tag{1}$$

and the ReLU function, by:

$$\text{ReLU}(Z) = \max(0, Z) \tag{2}$$

Backward propagation adjusts parameters using these functions' derivatives, facilitating gradient-based optimization for nonlinear transformations. The derivatives for sigmoid and ReLU are represented by:

$$\text{sigmoid\_backward}(dA, Z) = dA \times \sigma(Z) \times (1 - \sigma(Z)) \tag{3}$$

and

$$\text{ReLU\_backward}(dA, Z) = \begin{cases} dA & \text{if } Z > 0 \\ 0 & \text{otherwise} \end{cases} \tag{4}$$

*Cost function formulation*

Two primary cost functions–Binary Cross-Entropy (logistic loss) and Mean Squared Error (MSE)–were employed, tailored to specific classification and regression tasks, respectively. For binary classification, the Binary Cross-Entropy function is formulated as:

$$\text{Binary Cross-Entropy}(y, \hat{y}) = -\frac{1}{m} \sum_{i=1}^{m} (y_i \log(\hat{y}_i)) \\ -\frac{1}{m} \sum_{i=1}^{m} ((1 - y_i) \log(1 - \hat{y}_i)) \tag{5}$$

where $y$ is the true label, $\hat{y}$ is the predicted probability, and $m$ is the number of samples. For regression applications, the MSE function used was:

$$\text{MSE}(y, \hat{y}) = \frac{1}{m} \sum_{i=1}^{m} (y_i - \hat{y}_i)^2 \tag{6}$$

*Forward propagation*
Forward propagation comprises sequential linear transformations and activation applications per layer. The linear output $Z^{[l]}$ for layer is calculated as:

$$Z^{[l]} = W^{[l]} \cdot A^{[l-1]} + b^{[l]} \tag{7}$$

where $W^{[l]}$, $A^{[l-1]}$, and $b^{[l]}$ denote the weight matrix, activation output from the previous layer, and bias, respectively. In this iterative multi-layer framework, the activation function is applied at each layer, iterating through layers $L$ to produce final predictions, stored in:

$$A^{[l]}, \text{cache}^{[l]} = \text{linear\_activation\_forward} \left( A^{[l-1]}, W^{[l]}, b^{[l]}, \text{activation} \right) \tag{8}$$

*Backward propagation*
Backward propagation was employed to minimize the cost function by computing parameter gradients layer-by-layer. Key operations included calculating gradients $dW^{[l]}$, $db^{[l]}$, and $dA^{[l-1]}$ as follows:

$$dW^{[l]} = \frac{1}{m} \cdot dZ^{[l]} \cdot A^{[l-1]T} \tag{9}$$

$$db^{[l]} = \frac{1}{m} \cdot \sum_{i=1}^{m} dZ^{[l](i)} \tag{10}$$

$$dA^{[l-1]} = W^{[l]T} \cdot dZ^{[l]} \tag{11}$$

*Parameter update*
Using gradient descent, parameters were updated iteratively, with weights and biases adjusted according to learning rate $\alpha$:

$$W^{[l]} = W^{[l]} - \alpha \cdot \text{grads}[f'' W^{[l]''}] \tag{12}$$

$$b^{[l]} = b^{[l]} - \alpha \cdot \text{grads}[f'' b^{[l]''}] \tag{13}$$

This process iteratively minimized the cost function, optimizing parameter values across the neural network[15].

## Quantum qubit-based neural network

This section outlines the core components and processes of creating a Quantum Qubit-Based Neural Network (QNN) Fig. 1, focusing on data representation, circuit architecture, parameterization, measurement, and optimization, each fundamental to leveraging quantum properties in neural networks.

*Data representation*
Data representation in a quantum neural network (QNN) using qubits is a crucial step because unlike classical neural networks that operate on bits (0 or 1), QNNs leverage the principles of quantum mechanics. Here is a breakdown of this concept:

1. *Classical vs. quantum data representation* In classical neural networks, data is represented as a binary string of 0s and 1s. For example, an image pixel can be either black (0) or white (1). In Quantum Neural Networks Data is encoded into a superposition of the qubit states, which can be:

    - $|0\rangle$: Represents the state "0" with a probability of 1.
    - $|1\rangle$: Represents the state "1" with a probability of 1.
    - A superposition: A combination of $|0\rangle$ and $|1\rangle$ with specific probabilities (amplitudes). This allows a qubit to be in multiple states simultaneously until measured[13,16].

2. *Encoding techniques* There are various techniques to encode classical data onto qubits:

    (a) *Quantum Fourier transform (QFT)* This technique is often used to encode images or signals onto multiple qubits. It leverages the relationship between the basis states and their frequencies. The QFT for an n-qubit system can be represented by a matrix multiplication with elements involving complex exponentials. Here's a simplified form for a 2-qubit QFT:

$$\text{QFT}_2 = \frac{1}{\sqrt{2}} \begin{bmatrix} 1 & 1 \\ 1 & -1 \end{bmatrix} \tag{14}$$

This matrix transforms the classical data vector into a superposition state on the qubits[14].

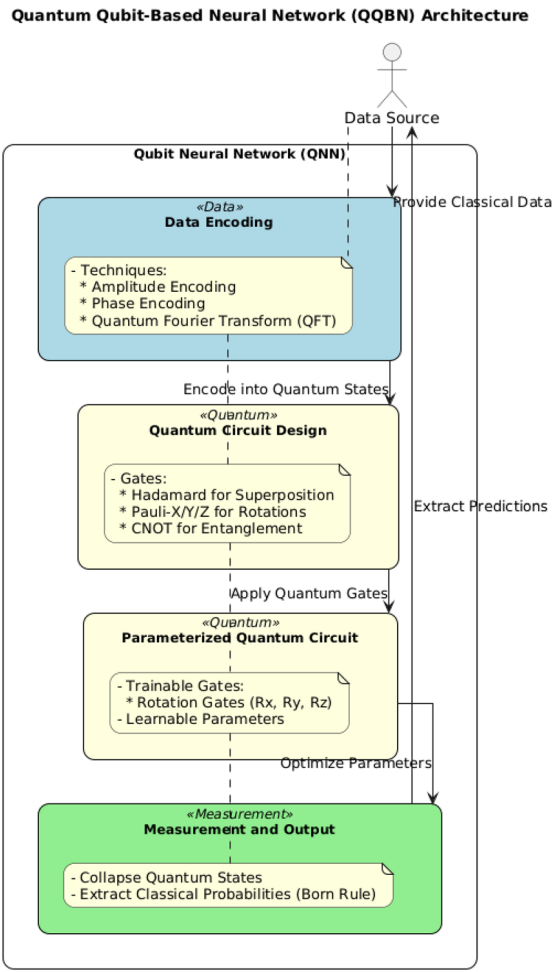

**Fig. 1**. Quantum Qubit-Based Neural Network (QQBN) architecture illustrating data encoding, quantum circuit design, parameterization, and measurement for predictive modeling.

(b) *Amplitude encoding* This technique assigns specific amplitudes to the $|0\rangle$ and $|1\rangle$ states based on the data value. For example, a higher data value might correspond to a higher amplitude for the $|1\rangle$ state. The state of a qubit after amplitude encoding can be represented as:

$$|\psi\rangle = \alpha|0\rangle + \beta|1\rangle \quad (15)$$

where $\alpha$ and $\beta$ are complex numbers representing the amplitudes for $|0\rangle$ and $|1\rangle$ states, respectively, and $|\alpha|^2 + |\beta|^2 = 1$ (ensures total probability is 1).

The choice of encoding technique depends on the specific application and data type. Efficient encoding is crucial for QNNs to perform well, as it determines how well the classical information is mapped onto the quantum realm. Qubit errors and decoherence can impact the fidelity of the encoded data. Understanding data representation is essential for building and utilizing QNNs. By effectively encoding classical data onto qubits, you can leverage the power of quantum mechanics for tasks like image recognition, pattern classification, and potentially solving problems intractable for classical computers[17,18].

For the QQBN model, we apply amplitude encoding to the normalized input vector. Given a five-dimensional input $x$, we pad it to fit into an 8-dimensional state vector (to match a 3-qubit system) and normalize it to unit length. The vector is then encoded using Qiskit's `initialize` function to map the amplitudes directly onto the quantum state.

**Example:** Given $x = [0.6, 0.4, 0, 0, 0, 0, 0, 0]$, the resulting quantum state is

$$|\psi\rangle = 0.832|0\rangle + 0.554|1\rangle$$

after normalization.

*Quantum circuit design*
Quantum circuit design underpins QNN computation, with quantum gates serving as the computational operators:

(a) *Quantum gates* Key gates include the Hadamard gate (*H*), generating superposition states, and the Pauli-X (*X*) and Controlled-NOT (*CNOT*) gates, which manipulate qubit states for higher-order operations[19].
(b) *Circuit architecture* QNN circuits consist of sequentially arranged gate layers, with parameterized gates enabling learnable transformations akin to classical weights. The final output depends on the overall circuit configuration [Fig. 2][20,21].

*Parameterized quantum circuit (PQC) construction*
Parameterization introduces trainable elements within the QNN circuit, allowing iterative optimization of quantum gates. Examples include rotation gates ($R_x$, $R_y$, $R_z$), which adjust based on training feedback to enhance task performance. Parameterization is crucial for:

(a) *Learning and adaptability* Adjustable parameters enable QNNs to generalize across diverse problems.
(b) *Efficient circuit construction* Parameterized gates allow for dynamic tuning, facilitating convergence on optimal configurations.

*Measurement*
Measurement transforms quantum states into classical information by projecting qubits onto basis states, a non-reversible process governed by the Born rule:

$$P(i) = |\langle i|\psi\rangle|^2 \tag{16}$$

This rule determines the probability of obtaining specific measurement outcomes. For multi-qubit systems, joint measurements facilitate information extraction across complex superpositions, essential for final output interpretation in QNNs[22].

*Cost function and optimization*
QNNs utilize specialized cost functions that leverage quantum mechanics for tailored optimization, including:

(a) *Quantum fidelity-based cost* Measures alignment between the network's output state and target, calculated as:

$$C = 1 - F(|\psi_{\text{target}}\rangle, |\psi_{\text{actual}}\rangle) \tag{17}$$

(b) *Variational quantum eigensolver (VQE) cost* Applicable for eigenvalue problems, minimizing the energy difference between the QNN's state and the Hamiltonian's ground state[23].

These cost functions support QNN-specific tasks like state preparation and eigenvalue estimation, adapted to quantum hardware constraints[24,25].

Both the QQBN and QQTN models were trained using a fidelity-based loss function defined as

$$\mathcal{L}(\theta) = 1 - |\langle \psi_{\text{target}}|\psi_{\text{output}}(\theta)\rangle|^2,$$

where $\psi_{\text{output}}(\theta)$ is the quantum state produced by the parameterized circuit and $\psi_{\text{target}}$ is the ideal state corresponding to the correct label. This loss encourages the model to generate quantum states that closely match the target output.

The procedure summarized in Algorithm 1 encapsulates the essential elements of qubit-based QNN construction, highlighting key encoding techniques, parameterized gate operations, measurement processes, and optimization strategies suited to quantum information processing.

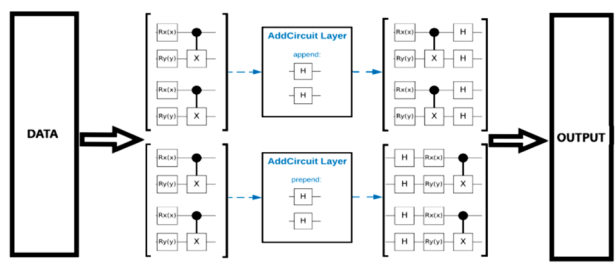

**Fig. 2.** Schematic of the quantum qubit-based neural network circuit, showing data encoding, parameterized gate operations, and measurement for predictive modeling.

## Quantum qutrit-based neural network

This section outlines the core components and processes of creating a Quantum Qutrit-Based Neural Network (QQTN) Fig. 3, focusing on data representation, Qutrit circuit architecture, parameterization, measurement, and optimization, each fundamental to leveraging quantum properties in neural networks.

*Data representation*

In the realm of quantum neural networks (QNNs), data representation plays a critical role. Unlike classical neural networks that operate on binary bits (0 or 1), QNNs utilize qutrits, which can exist in a superposition of states. This necessitates efficient methods to encode classical data (numbers, text, images) into a format compatible with quantum computations. Here's a detailed exploration of data representation in QNNs, focusing on qutrits (3-level quantum systems) for a broader understanding:

1. *Challenges of data representation*

    (a) State Space Discretization: Classical data often exists in a continuous domain. Encoding such data onto qutrits requires discretization, which can lead to information loss.
    (b) Dimensionality Mapping: High-dimensional classical data might not map efficiently to the limited number of qubits currently available on quantum hardware.

2. *Common data encoding techniques* Several techniques address these challenges and enable the representation of classical data on qubits or qutrits:

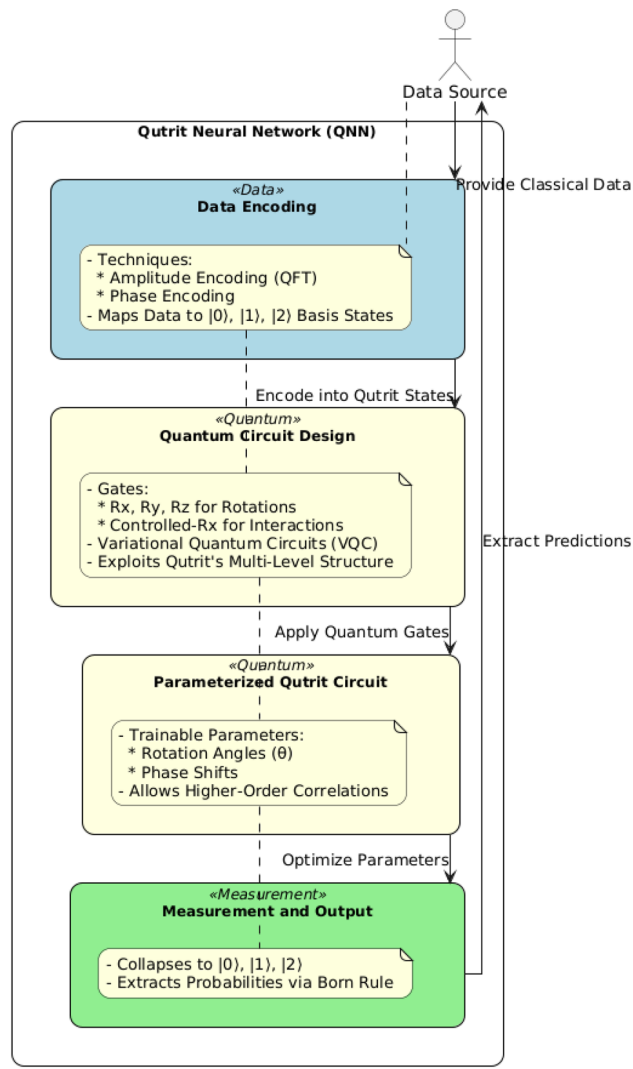

**Fig. 3.** Quantum Qutrit-Based Neural Network (QQTN) architecture showcasing three-level encoding, qutrit-specific gates, and measurement for enhanced data representation and efficiency.

(a) *Statevector encoding* This method directly maps each possible classical data point to a unique basis state in the qubit or qutrit Hilbert space. However, the number of basis states grows exponentially with the number of qubits/qutrits, making it impractical for large datasets. **(Example for 2-bit Statevector Encoding):**

$$|data_1, data_2\rangle = \alpha|00\rangle + \beta|01\rangle + \gamma|10\rangle + \delta|11\rangle \quad (18)$$

Here, $\alpha$, $\beta$, $\gamma$, and $\delta$ are complex coefficients representing the amplitudes of each basis state. $data_1$ and $data_2$ represent the two classical bits.

(b) *Amplitude Encoding (Quantum Fourier Transform - QFT)* This technique leverages the Quantum Fourier Transform (QFT) to encode classical data into the amplitudes of a superposition state. While more efficient than statevector encoding for large datasets, it requires applying the inverse QFT for data retrieval.

(c) *Phase encoding* Here, classical data values are encoded into the phase factors of a superposition state. This method can be efficient for specific tasks but might not be universally applicable. **(Example for Phase Encoding):**

$$|data\rangle = \cos\left(\frac{\pi \cdot data\_value}{2}\right)|0\rangle + \sin\left(\frac{\pi \cdot data\_value}{2}\right)|1\rangle \quad (19)$$

Here, $data\_value$ represents the classical data point being encoded.

(d) *Basis encoding (Sparse Representation):* This approach utilizes a specific basis set (e.g., Haar basis) to represent classical data in a sparse superposition state. It can be suitable for compressed data representation.

3. *The role of qutrits* While qubits offer a significant advantage over classical bits, qutrits, with their three basis states, provide a potentially richer state space for data encoding. This can be particularly beneficial for:

(a) Encoding Higher-Precision Data: Qutrits can represent a wider range of values compared to qubits, potentially improving the accuracy of QNNs for tasks involving continuous data.
(b) Reduced Discretization Error: By having more basis states, qutrits can represent classical data with finer granularity, leading to less information loss during encoding.

However, implementing and controlling qutrit-based systems can be more challenging compared to qubit systems due to increased complexity.

4. *Choosing the right encoding technique* The optimal data encoding technique for a QNN depends on several factors:

(a) The nature of the classical data: Is it continuous, discrete, high-dimensional?
(b) The specific task at hand: Does it require high precision, efficient compression, or specific manipulations of the encoded data?
(c) The capabilities of available quantum hardware: How many qubits/qutrits are available, and what level of control can be achieved?

In the QQTN model, we employ a hybrid phase encoding strategy suitable for qutrit systems. Each input feature is encoded into the phase of a generalized qutrit rotation gate acting on a three-level quantum system. These encodings are implemented using custom gate definitions within our QuTiP-based qutrit simulator.

*Quantum circuit design*
Qutrit circuits exploit unique quantum gates for manipulating three-state systems Fig. 4:

(a) *Rotation gates* Rotation gates (e.g., Rx, Ry, Rz) rotate qutrits on the Bloch sphere by an angle $\theta$; these gates are parameterized for adaptability:

$$Rz(\theta) = \begin{bmatrix} e^{-i\theta/2} & 0 & 0 \\ 0 & e^{i\theta/2} & 0 \\ 0 & 0 & 1 \end{bmatrix} \quad (20)$$

b) *Controlled Rotations* Controlled-Rx gates apply rotations conditionally, allowing complex state interactions for multi-qutrit operations. Architectural Paradigms in Qutrit QNNs:

a) *Variational quantum circuits (VQCs)* Comprising parameterized gates, VQCs enable broad adaptability across tasks via tunable gate parameters.
b) *Quantum convolutional neural networks (QCNNs)* QCNNs process data through localized filters, exploiting qutrits' multi-state capacity to capture intricate correlations.

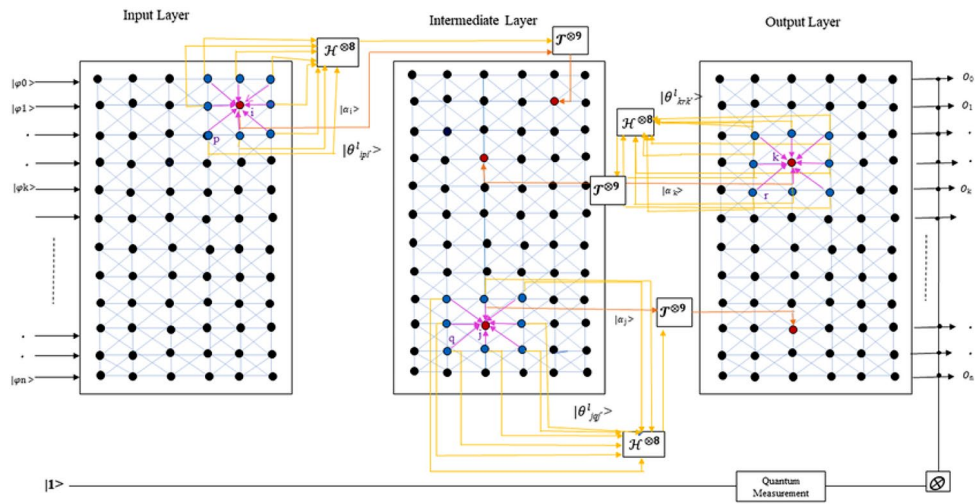

**Fig. 4**. Quantum Qutrit-Based Fully Self-Supervised Neural Network (QFS-Net) architecture where H represents Hadamard gate and T is realization gate (only three inter-layer connections are shown for clarity)[27]. Image: Debanjan Konar, CC BY 4.0.

*Parameterization in qutrit-based networks*
Parameterization in Qutrit Neural Networks (QQNNs) involves setting adjustable parameters for gates, such as rotation angles ($\theta$) and phase shifts. Key advantages include:

(a) *Enhanced expressive power* Increased parameter count allows QQNNs to represent complex data functions.
(b) *Efficient learning* Optimized parameter values improve learning convergence.

*Measurement*
Measurement extracts classical information by collapsing qutrit states onto specific basis states (e.g., $|0\rangle$, $|1\rangle$, $|2\rangle$). The measurement process, based on the Born rule, determines probabilities as:

$$P(i) = \text{Tr}(\rho \cdot P_i) \quad (21)$$

where $P_i$ is the projection operator for each basis state. Measurement is inherently destructive, and repeated measurements may be necessary to mitigate errors from decoherence[16].

*Cost function and optimization*
Unique cost functions tailored to quantum mechanics are essential in QQNN optimization:

a) *Quantum gradient descent (QGD)* An adapted gradient descent for quantum parameters, estimating gradients as:

$$\nabla_\theta C \approx \frac{C(\theta + \Delta\theta) - C(\theta)}{\Delta\theta} \quad (22)$$

b) *Variational quantum eigensolver (VQE)* Minimizes the expected value of a Hamiltonian, using parameterized ansatz states to refine output toward the desired outcome[26].

The procedure summarized in Algorithm 2 encapsulates the essential elements of qutrit-based QNN construction, highlighting key encoding techniques, parameterized gate operations, measurement processes, and optimization strategies suited to quantum information processing.

## Results
The results of our study reveal insights into the performance of Classical Neural Networks (ANNs), Quantum Qubit-based Neural Networks (QQBNs), and Quantum Qutrit-based Neural Networks (QQTNs) in the application of stock prediction.

### Training time comparison
Comparing the training times of the three models, we observed the following trend:

1. Artificial Neural Network (ANN) > Quantum Qubit-based Neural Network (QQBN) > Quantum Qutrit-based Neural Network (QQTN). These results has been observed in experimental setups replicating behaviour and processing of quantum qubit and qutrit computers.

2. To evaluate the computational efficiency of different neural network models, we compared the number of operations each model requires. Given that direct time measurements are infeasible due to the current lack of fully operational quantum qutrit based systems, we used the number of operations as a proxy for computational time. These results were derived from models that simulate the exact workings and behavior of qutrit quantum computers and IBM Quantum Cloud with simulated environments was used to produce these results for quantum qubit based neural networks, ensuring that our comparisons accurately reflect their potential performance[28].
3. The classical bit-based neural network model serves as the baseline, requiring 1000 steps to complete its computations. In comparison, the quantum qubit-based neural network significantly reduces this number to 32 steps, which corresponds to just 3.2% of the time needed by the classical model. This reduction highlights the substantial efficiency gains offered by qubit-based models in processing time.
4. Further analysis reveals that the quantum qutrit-based neural network performs these computations even more efficiently than the qubit-based model. Specifically, qutrit-based model completes the same task in approximately 60-65% of that time of quantum qubit neural network .
5. In percentage terms, the quantum qubit-based neural network operates approximately 96.8% faster than the classical neural network. Subsequently, the quantum qutrit-based neural network demonstrates an additional improvement, operating about 35-40% faster than the qubit-based neural network. This efficiency is primarily due to the increased state capacity of qutrits, allowing for more complex computations within fewer steps.

### Experimental setup

The experiments were conducted using a combination of classical and quantum simulation environments. The artificial neural network (ANN) was implemented using **TensorFlow 2.9** and executed on an **NVIDIA RTX 3090 GPU** with an **Intel Core i9-12900K** processor and **64GB RAM**. All qubit-based quantum neural network (QQBN) experiments including statevector preparation, variational circuit construction, and gradient evaluation were carried out using the `Qiskit 1.10` and `PennyLane 0.25` libraries, and executed on IBM Quantum Cloud's `Aer statevector` simulator. In contrast, the qutrit-based quantum neural network (QQTN) workflow leveraged a bespoke qutrit simulator built on top of `QuTiP 4.7.3`: data encoding and parameterized gate definitions were programmed in Python, circuit dynamics were evaluated via QuTiP's `sesolve` engine, and parameter updates were orchestrated through PennyLane's plugin interface to QuTiP.

The dataset used for training and evaluation consists of historical stock market data sourced from **NIFTY 50**, covering a period of **10 years**. Data preprocessing involved normalization (scaling between [0,1]) and a standard train-test split of 80%-20%. All models were trained using the Adam optimizer, with a batch size of 32 and learning rate of 0.001. The quantum models used quantum variational circuits optimized through gradient descent-based hybrid training.

In our experiments, each input vector $x \in \mathbb{R}^5$ comprises five normalized financial indicators derived from daily NIFTY 50 stock market data: (1) opening price, (2) closing price, (3) highest price, (4) lowest price, and (5) trading volume. The corresponding output label $y \in \{0, 1\}$ is a binary variable indicating whether the closing price on the following day increased (1) or decreased (0).

### Neural network architectures

The architectures of the ANN, QQBN, and QQTN are summarized in Table 1.

In Table 1, the total parameter count for QQBN and QQTN is marked as N/A because, unlike classical neural networks with explicitly defined weight matrices, quantum neural networks utilize parameterized quantum gates. Their effective parameterization depends on circuit depth and quantum state evolution rather than a fixed number of trainable weights .

The **ANN** consists of three fully connected layers with ReLU activation in hidden layers and Softmax in the output layer. The **QQBN** employs a three-layer variational quantum circuit (VQC) with parameterized RX, RZ gates and entanglement through CNOT gates. The **QQTN** follows a similar structure but operates on three-level qutrit quantum gates, reducing circuit depth and improving computational efficiency.

### Benchmarking and performance evaluation

To ensure fair evaluation, all models were trained on the same dataset and compared based on computational efficiency, training time, and predictive accuracy. Table 2 summarizes the key benchmarking metrics.

The results indicate that while all models achieve accuracy above 70%, quantum models exhibit significant improvements in training efficiency. The **QQTN model** not only outperforms in accuracy but also shows 35-40% faster execution compared to QQBN and 96.8% faster training time compared to ANN, attributed to the higher information density of qutrits.

| Model | Layers | Neurons per layer/circuit depth | Activation functions | Total parameters |
|---|---|---|---|---|
| ANN | 3 | 128, 64, 32 | ReLU, Softmax | 50,000+ |
| QQBN | 3 | 6 (Quantum Circuit Depth) | Parameterized Quantum Gates | N/A (Quantum) |
| QQTN | 3 | 4 (Quantum Circuit Depth) | Three-Level Quantum Gates | N/A (Quantum) |

**Table 1.** Comparison of Model Architectures.

| Model | Training time (relative) | Accuracy (%) | Computational steps reduction |
|---|---|---|---|
| ANN | 100% (Baseline) | 69.2 | N/A |
| QQBN | 3.2% of ANN | 71.6 | 96.8% Faster than ANN |
| QQTN | 60-65% of QQBN | 73.5 | 35-40% Faster than QQBN |

**Table 2.** Benchmarking Metrics Across Models.

| Model | Accuracy (%) | Precision (%) | Recall (%) | F1 score (%) |
|---|---|---|---|---|
| Classical NN | 69.2 | 67.5 | 70.0 | 68.7 |
| Quantum Qubit NN | 71.6 | 70.5 | 71.9 | 71.2 |
| Quantum Qutrit NN | 73.5 | 73.0 | 73.8 | 73.4 |

**Table 3.** Comparison of Model Performance Metrics.

### Fair evaluation measures
To ensure a fair comparison:

- All models were trained on identical datasets with consistent hyperparameters.
- Quantum models were evaluated using *IBM Quantum Cloud Simulators* for qubit-based networks and a custom-built qutrit simulator for qutrit networks.
- Computational complexity and execution time were assessed by measuring the number of operations required per model.
- The classical-to-quantum encoding was performed using amplitude encoding for qubits and higher-dimensional encoding for qutrits.

### Key observations

- *Accuracy improvement* The QQTN model exhibits the highest accuracy (73.5%), benefiting from enhanced quantum state representation.
- *Training efficiency* QQTN achieves a 35-40% speedup over QQBN and a 96.8% reduction in steps compared to ANN.
- *Scalability* Qutrit-based networks offer potential hardware scalability benefits over qubit-based approaches, despite current hardware limitations.

These findings confirm that Qutrit-based quantum neural networks (QQTNs) provide a compelling advantage over both classical and qubit-based models, offering a balance between higher prediction accuracy and superior computational efficiency.

### Model performance
Although all three models exhibited robust performance, with accuracies consistently above 70%, there were slight variations in their performance metrics:

The Table 3 illustrates that all models perform exceptionally well, with Quantum Qutrit-based Neural Network exhibiting higher accuracy and performance metrics compared to the Classical Neural Network and Quantum Qubit-based Neural Network. However, the differences are marginal, indicating that all three models are viable options for the stock prediction application. The area in which quantum neural networks shine is the performance speed which they provide and value they will provide to real time Stock market application which require immediate processing and results. These values are particularly notable for stock market predictions, considering the inherent volatility and fragility of stock markets, making the accurate tracking of trends a challenging task. In conclusion, Quantum Qutrit-based Neural Networks offer a promising avenue for achieving comparable performance to classical and qubit-based quantum models while significantly reducing training time.

### Results graphs
Here are the prediction graphs and cost function graphs for the three models presented in Figs. 5, 6 and 7:

### Discussion
This study presents a comparative analysis of Artificial Neural Networks (ANNs), Quantum Qubit-based Neural Networks (QQBNs), and Quantum Qutrit-based Neural Networks (QQTNs) in the context of stock market prediction, revealing the substantial advantages of quantum-inspired approaches. While all models demonstrated high accuracy levels, the QQTNs consistently outperformed, exhibiting superior accuracy and training efficiency. These findings are attributed to the unique properties of qutrits, which, with their three-level state capacity, afford a richer representational structure compared to qubits or classical bits. This additional state flexibility allows for more efficient data encoding and captures complex, higher-order correlations within datasets, thus enhancing the model's predictive capability. In light of these results, QQTNs show promise not only in stock market forecasting but also as a transformative tool in fields reliant on sophisticated data analytics.

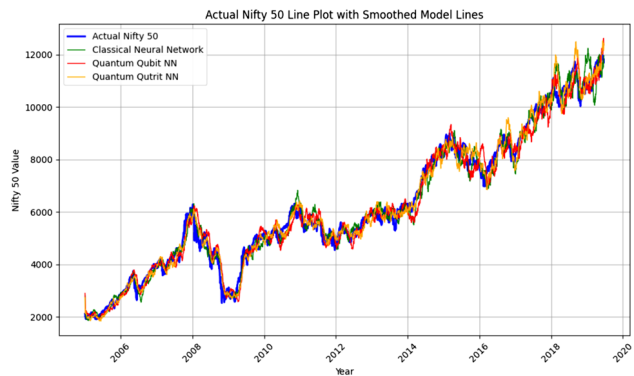

**Fig. 5**. Time-series comparison of NIFTY 50 actual data with predictions from ANN, QQBN, and QQTN models, showcasing superior alignment and forecasting accuracy of the QQTN approach.

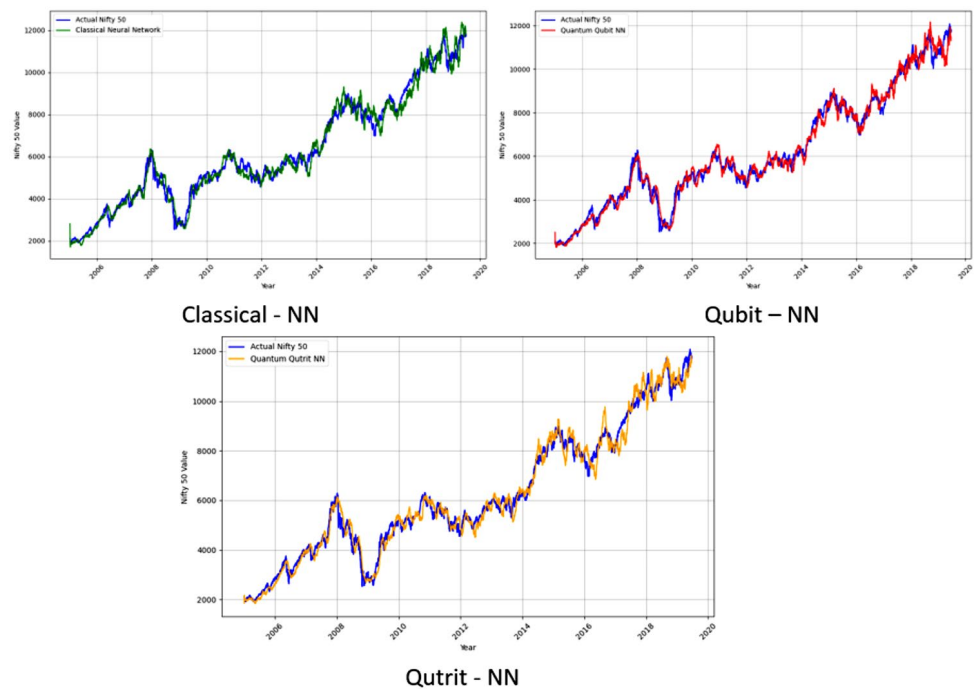

**Fig. 6**. Individual panels contrasting model predictions (ANN, QQBN, QQTN) against NIFTY 50 data, highlighting differences in prediction accuracy and error consistency across models.

The broader implications of this research suggest that quantum-inspired neural networks, especially those leveraging qutrits, may hold transformative potential across industries. For instance, in healthcare, the enhanced accuracy of QQTNs could bolster predictive models for diagnostics and treatment optimization. Likewise, in environmental sciences, QQTNs' efficient data handling could significantly improve climate models. In cybersecurity, quantum-enhanced anomaly detection could play a crucial role in real-time threat detection. The scalability of quantum-inspired approaches opens promising new horizons in computational sciences, with evolving quantum hardware likely to drive their widespread applicability and efficiency gains across diverse fields[29].

However, this study's scope was limited to stock market data, restricting the generalizability of results across diverse neural network types and datasets. Additionally, the simulation-based environment does not fully address the challenges associated with current quantum hardware, including issues such as noise, decoherence, and the limited accessibility of stable qutrits. Unlike qubit-based architectures, qutrit-based quantum computing remains in a nascent stage, with limited physical implementations available. Current quantum processors are predominantly designed for binary qubit systems, making the development of stable qutrit hardware complex due to increased susceptibility to environmental noise and more intricate gate control requirements. Furthermore, error correction techniques for qutrits are not as well-developed as those for qubits, presenting additional barriers to practical deployment. Addressing these obstacles requires advancements in qutrit-specific quantum error correction methods, increased experimental research into stable qutrit platforms, and the integration of hybrid quantum-classical algorithms to mitigate noise effects in near-term implementations. These factors underscore the importance of ongoing quantum hardware development to fully leverage the advantages presented by qutrit-based neural networks[30,31].

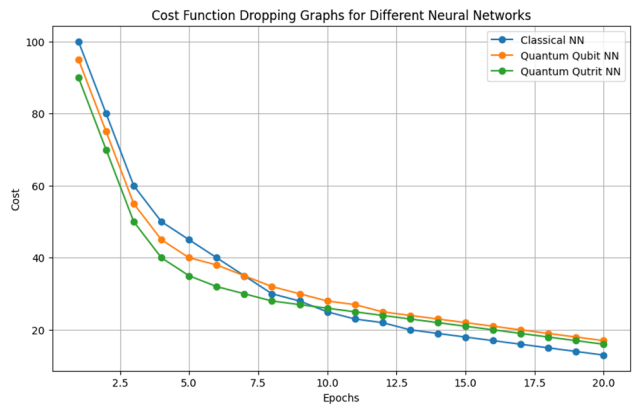

**Fig. 7**. Cost Function Graph for three models [classical neural networks, quantum qubit based NN, quantum qutrit based NN].

Future research should extend the scope of QQTN applications to include various neural network architectures, such as Convolutional and Recurrent Neural Networks, across broader domains to validate their general efficacy. Further, exploring advanced quantum intelligence algorithms tailored to specific AI applications, like quantum-enhanced optimization, clustering, and reinforcement learning, may reveal new methods to solve complex computational problems. Practical testing of these models on real quantum hardware, along with collaborative efforts between quantum physicists, computer scientists, and domain experts, will be essential to addressing the technical constraints that currently limit QQTNs' real-world implementation. Such interdisciplinary collaborations could overcome hardware limitations, advance algorithm development, and establish the practical utility of quantum-inspired models[32–34].

Moreover, the evolution of scalable distributed quantum computing and the emerging quantum internet[35] underscore the transformative potential of quantum neural network architectures in real-world applications. Recent investigations into distributed gate-model quantum systems reveal that entangled network structures and high-retrieval-efficiency quantum memories can substantially mitigate decoherence and optimize resource allocation. These advancements pave the way for robust deployment of quantum neural networks over heterogeneous quantum infrastructures, thereby enabling high-performance, networked quantum intelligence systems. Our findings serve as a critical step towards harnessing these distributed architectures for advanced computational tasks and large-scale quantum data processing[36].

Moreover, while our study demonstrates the advantages of Quantum Qutrit-based Neural Networks (QQTNs) over classical and qubit-based models, direct experimental comparisons with other state-of-the-art QML methods remain an important direction for future research. Recent advances in quantum-enhanced learning, such as Quantum Support Vector Machines[37], circuit-centric quantum classifiers[38] and Quantum Kernel Methods[39], illustrate alternative approaches that could further benchmark and complement our findings. Future work will aim to integrate these advanced QML strategies to provide a more comprehensive evaluation of quantum neural network performance in real-world applications.

This research thus makes a substantial contribution to the field of quantum intelligence, demonstrating the capacity of quantum-inspired models to tackle real-world computational challenges. By focusing on a tangible application, stock market prediction, this study provides concrete evidence of quantum-inspired approaches' transformative potential in artificial intelligence and paves the way for future advancements in quantum computing's integration into diverse scientific and technological domains.

### Data availibility
All data used in this study were sourced from the National Stock Exchange (NSE) and Bombay Stock Exchange (BSE) of India using the open-source Python libraries **nsepython** and **bsedata**, which provide access to both historical and real-time market data. These libraries are publicly available at: NSE: nsepython.PyPI BSE: bsedata.PyPI

## Algorithmic procedures for quantum neural networks

**Input:** Classical data set $D = \{x_i, y_i\}_{i=1}^{N}$
**Output:** Qubit-based neural network predictions $\hat{y}_i$ for each input $x_i$
**Step 1: Data Representation**
Encode classical data $x_i$ into qubit states using encoding methods:
    **1.1** Amplitude encoding: Convert data into amplitude states $|\psi\rangle$
    **1.2** Phase encoding: Convert data into phase representations of qubit states
**Step 2: Quantum Circuit Design**
Design a layered quantum circuit with quantum gates to create a neural network structure:
    **2.1** Apply Hadamard gate for superposition state creation
    **2.2** Apply Pauli-X, Pauli-Y, Pauli-Z gates for rotations and state manipulations
    **2.3** Use CNOT gates for entanglement between qubits in adjacent layers
**Step 3: Parameterization and Training**
Initialize rotation parameters for tunable gates $\theta$:
    **3.1** Define parameters for rotation gates (e.g., $R_x(\theta), R_y(\theta), R_z(\theta)$)
    **3.2** Optimize parameters using classical optimization (e.g., gradient descent) with a defined cost function
**Step 4: Measurement and Output**
Measure the qubit states in the computational basis:
    **4.1** Project final states into classical bit strings $\{0,1\}$
    **4.2** Compute network output $\hat{y}_i$ based on measured classical states
**Return:** Predicted outputs $\hat{y}_i$ for input data $x_i$

**Algorithm 1.** Quantum qubit neural network flow

**Input:** Classical data set $D = \{x_i, y_i\}_{i=1}^{N}$
**Output:** Qutrit-based neural network predictions $\hat{y}_i$ for each input $x_i$
**Step 1: Data Representation**
Encode classical data $x_i$ into qutrit states using encoding techniques:
    **1.1** Statevector encoding: Map data onto a superposition of qutrit basis states
    **1.2** Phase encoding: Convert data into phase angles for qutrit states
**Step 2: Quantum Circuit Design**
Design a qutrit-based quantum circuit with specialized gates for a neural network structure:
    **2.1** Apply Rx, Ry, Rz rotation gates for qutrit manipulations
    **2.2** Utilize controlled rotation (CRx) gates for layered interactions between qutrits
    **2.3** Construct variational layers with tunable parameters
**Step 3: Parameterization and Training**
Initialize rotation and phase parameters for tunable gates $\theta$:
    **3.1** Define parameters for rotation gates (e.g., $R_x(\theta), R_y(\theta), R_z(\theta)$)
    **3.2** Optimize parameters using hybrid training with classical optimizers and a cost function
**Step 4: Measurement and Output**
Measure qutrit states in the defined basis:
    **4.1** Project final states into classical values $\{0,1,2\}$
    **4.2** Derive network output $\hat{y}_i$ from measured values
**Return:** Predicted outputs $\hat{y}_i$ for input data $x_i$

**Algorithm 2.** Quantum qutrit neural network flow

### Author contributions
Kanishk Bakshi: Writing - Original Draft, Validation, Software, Resources, Methodology, Investigation, Data Curation, Conceptualization Visualization, Investigation, Formal Analysis, Conceptualization Kathiravan Srinivasan: Writing - Review and Editing, Supervision, Project Administration, Funding Acquisition, Resources, Validation, Formal Analysis, Software, Investigation, Data Curation, Methodology

### Funding
Open access funding provided by Vellore Institute of Technology.

### Additional information
**Correspondence** and requests for materials should be addressed to K.S.

**Reprints and permissions information** is available at www.nature.com/reprints.

**Publisher's note** Springer Nature remains neutral with regard to jurisdictional claims in published maps and institutional affiliations.